\def\year{2020}\relax
\documentclass[letterpaper]{article} 
\usepackage{aaai20}  
\usepackage{times}  
\usepackage{helvet} 
\usepackage{courier}  
\usepackage[hyphens]{url}  
\usepackage{graphicx} 
\usepackage{graphicx}  
\usepackage{hyperref}  
\usepackage{amsfonts}
\usepackage{amsmath}
\usepackage{amsbsy}
\usepackage[]{algorithm2e}

\newcommand{\citet}[1]{\citeauthor{#1} \shortcite{#1}}
\newcommand{\citep}{\cite}

\newcommand{\diag}{\textrm{diag}}
\DeclareRobustCommand{\E}[2]{{\mathbb{E}_{#1}} \left[#2\right]}

\newcommand{\bone}{\mathbf{1}}
\newcommand{\bzero}{\mathbf{0}}
\newcommand{\ba}{\mathbf{a}}
\newcommand{\bb}{\mathbf{b}}
\newcommand{\bc}{\mathbf{c}}

\newcommand{\be}{\mathbf{e}}

\newcommand{\bg}{\mathbf{g}}
\newcommand{\bh}{\mathbf{h}}

\newcommand{\bs}{\mathbf{s}}

\newcommand{\bu}{\mathbf{u}}
\newcommand{\bv}{\mathbf{v}}
\newcommand{\bw}{\mathbf{w}}
\newcommand{\bx}{\mathbf{x}}
\newcommand{\by}{\mathbf{y}}

\newcommand{\bA}{\mathbf{A}}
\newcommand{\bB}{\mathbf{B}}

\newcommand{\bD}{\mathbf{D}}

\newcommand{\bH}{\mathbf{H}}
\newcommand{\bI}{\mathbf{I}}

\newcommand{\bM}{\mathbf{M}}

\newcommand{\bP}{\mathbf{P}}

\newcommand{\bS}{\mathbf{S}}
\newcommand{\bT}{\mathbf{T}}
\newcommand{\bU}{\mathbf{U}}

\newcommand{\bW}{\mathbf{W}}
\newcommand{\bX}{\mathbf{X}}

\newcommand{\bbeta}{\boldsymbol{\beta}}

\newcommand{\blambda}{\boldsymbol{\lambda}}
\newcommand{\bgamma}{\boldsymbol{\gamma}}

\newcommand{\bmu}{\boldsymbol{\mu}}

\newcommand{\bGamma}{\mathbf{\Gamma}}
\newcommand{\bLambda}{\mathbf{\Lambda}}

\newcommand{\bbN}{\mathbb{N}}

\newcommand{\bbR}{\mathbb{R}}

\title{Kriging Convolutional Networks}
\author{\Large \textbf{Gabriel Appleby\thanks{Equal contribution. The first two authors are arranged by the alphabet order}, Linfeng Liu\footnotemark[1] , Li-Ping Liu}\\ 
Department of Computer Science, Tufts University\\ 
\{Gabriel.Appleby, Linfeng.Liu, Liping.Liu\}@tufts.edu 
}
 
\begin{document}

\maketitle

\begin{abstract}
Spatial interpolation is a class of estimation problems where locations with known values are used to estimate values at other locations, with an emphasis on harnessing spatial locality and trends. Traditional Kriging methods have strong Gaussian assumptions, and as a result, often fail to capture complexities within the data. Inspired by the recent progress of graph neural networks, we introduce Kriging Convolutional Networks (KCN), a method of combining advantages of Graph Convolutional Networks (GCN) and Kriging. Compared to standard GCNs, KCNs make direct use of neighboring observations when generating predictions. KCNs also contain the Kriging method as a specific configuration. We further improve the model's performance by adding attention. Empirically, we show that this model outperforms GCNs and Kriging in several applications. The implementation of KCN using PyTorch is publicized at the GitHub repository\footnote{This is a new implementation updated in 2023 using PyTorch, so the performation values are slightly different with those reported in the paper.}: \url{https://github.com/tufts-ml/kcn-torch}.
\end{abstract}


\noindent Spatial data is ubiquitous in a wide variety of fields such as ecology \citep{fink2010}, economics \citep{gao2014data}, and meteorology \citep{xingjian2015convolutional}. A common task within these fields is to estimate values at target locations from nearby known values. Improving these estimations should provide clear benefits for these applications. Estimation techniques tailored to spatial data must leverage the fact that every data point is associated with a location. Most importantly, these techniques should be able to capture the spatial correlation among these locations.

In many fields, the most prevalent method for spatial data modeling is kriging \citep{cressie1991statistics}. The fundamental assumption of kriging is that observations at locations are from an underlying Gaussian process. After estimating the \textit{variogram}, which is essentially the strength of spatial correlations between data points, kriging uses a \textit{linear interpolation of observed values} to predict the value at a new location. The kriging prediction is the best linear unbiased estimator for spatial points given its Gaussian assumption. However, this assumption is quite constrictive, as data in many applications are not from a Gaussian distribution. For example, we will show in our experiments that this assumption leads to poor performance when estimating integer counts that contain a significant fraction of zeros.

Researchers also use flexible machine learning algorithms \citep{hengl2018random} for spatial data modeling. Given the huge success of GNNs and the similarity between spatial data and graph data, researchers have started to apply Graph Neural Networks (GNN) \cite{wu2019comprehensive} to spatial data \cite{li2017diffusion,yu2017spatio,zhu2018modelling,yan2019graph}. GNNs were first developed for explicit graph data, but can model any data that can be transformed into a graph either by their spatial vicinity or their physical connections (e.g. routes). The main idea is to propagate information along graph edges, so graph nodes can share information during the learning process. GNNs are relatively generic, and can find nonlinear relationships between the inputs, hidden layers, and neighborhood information of each node. By design, GNNs are more flexible than kriging. 

However, kriging has an advantage over GNNs: kriging directly uses observed training labels to predict the label of a new data point. In comparison, there is no straightforward way to feed training labels as input to a GNN. It is not feasible to directly feed training labels as part of the input because the GNN will directly output the given label of a training data point and learn nothing. Furthermore, spatial data modeling requires inductive learning -- the model needs to be able to make predictions for new locations that are not in the graph formed by training data. While kriging is intrinsically inductive, only a few GNNs such as GraphSAGE \citep{hamilton2017inductive} can work inductively.

Inspired by these two observations, we develop a new model, the Kriging Convolutional Network (KCN), as an improvement to GNNs. The KCN is still a type of GNN. However, it does not form a single large graph over all data points. Every time a KCN fits the label of a data point (call it the \textit{center}), it forms a small graph over the center and its neighboring \textit{training} data points. These neighbors are the $K$ nearest neighbors according to a distance metric. In the input to the KCN, we hide the label of the center node. The input consists of feature vectors for all nodes in the graph ($K+1$ nodes), as well as the labels of the \textit{neighbors}. The KCN also needs the adjacency matrix of the graph, which is defined to be the spatial kernel matrix or a normalized version of that. The target value of the KCN is the label of the center node.  We iterate over all of the training data, treating each node as the center to train the KCN model. The KCN uses the same structure to predict the label of a new data point.   

The KCN combines the best parts of both models. In comparison to the GNN, it is able to directly leverage training labels in prediction, and no re-training is necessary when new data points are introduced. In contrast to kriging, the KCN is more versatile. On a large dataset where overfiting is not an issue, the KCN has a clear advantage over kriging. Even though the KCN's underlying mechanisms are very different from kriging, our theoretical analysis reveals a deep connection between the two models. In fact, with a special configuration, the KCN can emulate kriging.

In summary, this work has three contributions: 
\begin{itemize} 
\item the development of the KCN, which is a GNN that directly uses training labels for prediction;   
\item the theoretical result showing that the KCN approximately recovers local universal kriging; and  
\item empirical studies indicating the KCN's advantage over baseline models.  
\end{itemize}


\section{Related Work}

Kriging \citep{cressie1991statistics} has been widely used in spatial data modeling. Using Kriging to model non-Gaussian data is often accomplished through careful transformation of labels \citep{saito2000geostatistical}. However, it is not always feasible to transform a variable to be Gaussian \citep{dance2018comparison}. One direction of exploration is to weaken the Gaussian assumption of kriging models \citep{wallin2015geostatistical}, but these methods are often specially designed for their respective applications.

GNNs are neural networks that work on graph data \citep{gori2005new}. \citet{wu2019comprehensive} and \citet{zhou2018graph} have done extensive surveys of this topic.  A GNN typically consists of a few layers, each of which has   
 a non-linear transformation of the hidden vectors and a step of information propagation between nodes. GNN architectures differ 
by how they propagate information among graph nodes \citep{kipf2016,atwood2016diffusion,hamilton2017inductive,velickovic2018}. When a GNN is applied to spatial data \citep{wu2019comprehensive,yan2019graph}, one first builds a graph over data points in the spatial area and then runs the GNN on the graph. To the best of our knowledge, all of these methods feed features as the input and fit labels by the output of the network. In this work, we develop our KCN model based on the Graph Convolutional Network (GCN) \citep{kipf2016} and Graph Attention Network (GAT) \citep{velickovic2018}.


\section{Background}

Suppose there are $N$ spatial data points, $(\bs, \bX, \by) = (s_i, \bx_i, y_i)_{i=1}^{N}$, where $s_i$, $\bx_i$, and $y_i$ are respectively the location, the feature vector, and the label of data point $i$. Usually a location $s_i$ is a GPS coordinate, $s_i \in \bbR^2$. There are $d$ features in a feature vector $\bx_i \in \bbR^d$. The domain of the target value $y_i$ is application-dependent. For example, $y_i \in \bbN$ when $y_i$ is a count, and $y_i \in \bbR^+$ when $y_i$ represents the precipitation level. One important task of spatial data modeling is to predict or estimate the value $y_*$ for a new location $s_*$ with a feature vector $\bx_*$. Let $\hat{y}_*$ denote the prediction.
 
\subsection{Kriging}

There are many variants of kriging, of which \emph{universal kriging} is the most appropriate for the setting above. Universal kriging has the 
following model assumption (Eq. 3.4.2 in \citep{cressie1991statistics}).
\begin{align}
y_i = \bbeta^{\top} \bx_i + \epsilon(s_i), i = 1, \ldots, n, *
\end{align}
Here $\bbeta$ is the coefficient vector. $\epsilon(\cdot)$ is a zero-mean random process with \emph{variogram} $2\gamma(\cdot)$. The variogram $2\gamma(\cdot)$, which specifies the spatial correlation between data points,  is a function of spatial distance: $2\gamma(\|s_i - s_j \|)= \E{}{(\epsilon(s_i) - \epsilon(s_j))^2}$. The variogram often takes a special function form with its parameters estimated from the data. With this model assumption, kriging minimizes the expected squared error, $\E{y_{*}}{(\hat{y}_{*} - y_{*})^{2}}$, in closed form. Then the prediction $y_{*}$ of universal kriging is $\hat{y}_{*}^{kriging} =  \blambda^\top (\by - \bmu)$ with 
\begin{align}
\blambda = \bGamma^{-1} \left(\bgamma - \bB \bX^\top\bGamma^{-1} \bgamma + \bB \bx_*,  \right), 
\label{eq:lambda}
\end{align}
with 
$\bB = \bX (\bX^\top \bGamma^{-1} \bX)^{-1}$, $\bGamma = \left[\gamma(\|s_i - s_j\|) \right]_{i,j=1}^n$, and $\bgamma = \left[\gamma(\|s_i - s_*\|) \right]_{i = 1}^n$.  

Note that kriging uses known training labels as well as all features as the input to make the prediction. Despite its complex form, kriging has a subtle relation with the KCN model proposed later.

\subsection{Graph Convolutional Networks}

Suppose we have a graph $G = (V, E)$, where $V = \{1, \ldots, M\}$ is set of data points, and $E$ is the edge set. Each data point $i \in V$ has a feature vector $\bx_i$ and a label $y_i$. Later, we will collectively denote $(\tilde{\bX}, \tilde{\by})$ as a stack of all features and labels for notational convenience. Let $\bA$ denote the adjacency matrix of the graph, and $\bar{\bA}$ denote the normalized adjacency matrix, 
\begin{align}
\bar{\bA} = \bD^{-\frac{1}{2}}(\bA + \bI) \bD^{-\frac{1}{2}}, 
\end{align}
with $\bD = \diag(\bA \bone + \bone)$ being the degree matrix plus one. Then a GCN \citep{kipf2016} takes $\bar{\bA}$ and $\tilde{\bX}$ as the input and fits known labels in $\tilde{\by}$ as the target. The GCN consists of $L$ GCN layers. Each GCN layer $\ell$ takes an input $\bH_{\ell - 1} \in \mathbb{R}^{n \times d_{\ell - 1}}$ and outputs a matrix $\bH_{\ell} \in \mathbb{R}^{n \times d_{\ell}}$. The layer is parameterized by a matrix $\bW^{\ell}$ with size $d_{\ell - 1} \times d_{\ell}$. Formally, the GCN is defined by 
\begin{align}
\bH^0 &= \tilde{\bX}, \\
\bH^{\ell} &= \sigma\left( \bar{\bA} \bH^{\ell - 1} \bW^{\ell} \right), ~~~ \ell = 1, \ldots, L \label{eq:gcn}\\ 
\hat{\by} &= \bH^{L}.
\end{align}
Here $\sigma(\cdot)$ is a non-linear activation function. 

<<<<<<< HEAD
The GCN considers a semi-supervised task, in which only part of the labels $\tilde{\by}$ are observed. The GCN then defines its training loss based on the known labels, and aims to predict unknown labels. In our method, we will consider to predict one data point $\hat{y}_*$ at a time; thus $M=N+1$. To predict a scalar for a graph node, the last layer $\bH^{L}$ has only $d_{L}=1$ column, and its entry corresponding to the new data point $(s_*, x_*)$ is the prediction $\hat{y}_*$. In practice, a two-layer GCN with $L=2$ is often sufficient. 
=======
A GCN considers a semi-supervised task, in which only part of the labels $\tilde{\by}$ are observed. A GCN then defines its training loss based on the known labels, and aims to predict unknown labels. To apply a GCN to the previous task, we form a graph for $(\bs, s_*)$, and put all features to the graph nodes. When predicting a scalar for a graph node, the last layer $\bH^{L}$ has only $d_{L}=1$ column, and its entry corresponding to the new data point $(s_*, x_*)$ is the prediction $\hat{y}_*$.
>>>>>>> 1a47e4691e50e54382fde6fcb76d469264f9d92d


\subsection{Kriging Convolution Network}
In this work, we develop a new learning model that directly use training labels as the input for predictions. We call this model a Kriging Convolution Network (KCN). 

We will first demonstrate how a KCN will be used for prediction. Let's treat a KCN model as a function $KCN(\cdot;\theta)$ parameterized by $\theta$. When predicting the label of a new data point $(s_*, x_*)$, the model ideally should use all information available to make the prediction, that is, $\hat{y}_* = KCN(\bs, \bX, \by, s_*, \bx_*)$. However, it is not feasible to consider all of the training points for just one prediction. It is not necessary either, because data points far from $s_*$ often have little influence over $y_*$ in many spatial problems. Therefore, we use the $K$ nearest neighbors of the new data point as the input. Denote the index set of these neighbors as $\alpha_* \subset \{1, \ldots, N\}$, then the predictive function becomes
\begin{align}
\hat{y}_* = KCN(\bs_{\alpha_*}, \bX_{\alpha_*}, \by_{\alpha_*}, s_*, \bx_*). 
\end{align}

To train the model, we treat every training point $i$ as a test point and fit its training label $y_i$. The model's output, $\hat{y}_i$, is compared against the true label. The difference of the two is measured by some loss function $\mathrm{loss}(y_i, \hat{y}_i)$. The learning objective of the model is to minimize the summation of all training losses
\begin{align}
\min_{\theta} &\sum_{i=1}^{N} \mathrm{loss}(y_i, \hat{y}_i), \nonumber \\
\hat{y}_i &= KCN(\bs_{\alpha_i}, \bX_{\alpha_i}, \by_{\alpha_i}, s_i, \bx_i). \label{eq:objective}
\end{align}
Here $\alpha_i$ is the set of neighbors of $i$ in the training set.

Now we construct the network architecture of the KCN, i.e. the function $ KCN(\bs_{\alpha_i}, \bX_{\alpha_i}, \by_{\alpha_i}, s_i, \bx_i)$. Instead of using locations, $s_i$ and $\bs_{\alpha_i}$, as features, we define a complete graph over the data points $i$ and its neighbors and then use a GCN to construct the predictive model. Denote $\beta_i = \{i\} \cup \alpha_i$ as the set containing the data point $i$ and its neighbors. We first define a graph over $\beta_i$ by constructing its adjacency matrix $\bA$ from a Gaussian kernel, 
\begin{align}
A_{jk} = \exp\left(-\frac{1}{2 \phi^2} \|s_j - s_k\|^2_2 \right), ~~ \forall j, k \in \beta_i. \label{eq:adj}
\end{align}
Here $\phi$ is the kernel length, which is a hyperparameter. In this graph, the edge $(j, k)$ has a large weight when $j$ and $k$ are near each other and vice versa.



Next we define the feature input to the GCN. The input should include features, $\bx_i$ and $\bX_{\alpha_i}$, and neighboring labels $\by_{\alpha_i}$.  Incorporating this information into a matrix will require a bit of care. We place $\by_{\alpha_i}$ and a zero in place of $y_i$ into a vector with length $(K+1)$, so the model has no access to $y_i$. We also use an indicator vector $\be$ to indicate that the instance $i$ is the one to be predicted. Then the GCN input is expressed by a matrix  $\bH^0$ with size $(K + 1) \times (2 + d)$. 
\begin{align}
\bH^{0} = \left[
\begin{array}{ccc}
0 & 1 & \bx_i^\top \\ 
\by_{\alpha_i} & \bzero & \bX_{\alpha_i}
\end{array}
\right]. \label{eq:h0}
\end{align}
The locations $\bs_{\beta_i}$ can be included in the feature matrix $\bX$ as features if there is reason to suspect spatial trends. 

Then the KCN model is defined to be a GCN followed by a dense layer. The KCN is formally defined as
\begin{align}
\bH^{L} &= GCN(\bA, \bH^0),  \\
\hat{y}_i &= \sigma \left(\be^\top \bH^{L} \bw_{den}\right). \label{eq:pred}
\end{align} 
Here $\bA$ and $\bH^0$ are the adjacency matrix and the input feature matrix constructed from the neighborhood of $i$. Note that every data point $i$ gets its own $\bA$ and $\bH^0$, whose index $i$ is omitted for notational simplicity. The vector $\be$ is the indicator vector for $i$: it takes the first vector of $\bH^{L}$, corresponding to $i$, as the input to the dense layer. The dense layer allows for a final transformation of the data without interference from neighbors.

The KCN parameters are all weight matrices, $\theta = \{\bW^1, \ldots, \bW^L, \bw_{den}\}$. We train the KCN model by minimizing the loss in \eqref{eq:objective}. Then we can predict the label of a new data point using its features and neighbors in the training set.   
Algorithm \ref{kcn-alg} summarizes the training procedure of the KCN. 

\begin{algorithm}[t]
    \textbf{Input:} $(\bs, \bX, \by)$, $K$\

    \textbf{Output:} $\theta = (\bW^1, \ldots, \bW^L, \bw_{den})$ \

    \For{$i\gets0$ \KwTo $N$ }{
        $\beta_{i}$ = the $K$ nearest neighbors of $s_i$ and $i$ \;
        Compute $\bA$ from $s_{\alpha_{i}}$  by \eqref{eq:adj} \;
        Prepare $\bH^{0}$ from $(x_i, \bX_{\alpha_{i}}, y_{\alpha_{i}})$ by  \eqref{eq:h0} \;
    }

    \For{$iter \gets0$ \KwTo  num training iter}{
        $i = iter \% N $ \;
        $\bH^{L} = GCN(\bA, \bH^0; \bW^{1}, \ldots,  \bW^{L})$ \;
        $\hat{y}_i = \sigma (\be^\top \bH^{L}) \bw_{den}$ \;
        Compute $loss(y_i, \hat{y}_i)$ and its derivative \;
        Update weights $\theta = \bW^{1}, \ldots, \bW^{L}, \bw_{den}$
    }
    \caption{The training algorithm of KCN}
    \label{kcn-alg}
\end{algorithm}

Compared to local kriging, which only uses nearest neighbors for kriging, the KCN uses the same input. However, the KCN is much more flexible. When the training set is large enough such that the overfitting issue is less of a concern, the KCN model has clear advantages. 

Compared to the direct application of a GCN on spatial data, a KCN is able to use labels from neighbors directly. Furthermore, a KCN does not need to use the test data points to form the graph. Therefore, it does not need to re-train the model when there is a new batch of test data points.

The KCN is also similar to the KNN classifier but is much more powerful: while the KNN simply averages the labels of neighbors, the KCN uses a neural network as the predictive function.

\subsection{KCN with Graph Attention}

The recent success of attention mechanism on GNNs inspires us to try the Graph Attention network (GAT) \citep{velickovic2018} as the predicting model. The original GAT model computes attention weights with a neural network; it also requres that the attention weights of a node's neighbors sum up to 1. Here we use the dot-product self-attention \citep{vaswani2017} so that the model has a choice to fall back on the GCN model. 

Suppose the input feature at the $\ell$-th layer of the GCN is $\bH^{\ell - 1}$, then we compute an attention matrix $\bU$ by 
\begin{align}
\bP = \bH^{\ell-1}  \bW_{att}, \quad \bM = \sigma(\bP \bP^\top), \nonumber\\
\bLambda = \diag(\bM), \quad \bU = \bLambda^{-\frac{1}{2}} \bM  \bLambda^{-\frac{1}{2}}.
\end{align}
Here $\bW_{att}$ is the weight matrix for the attention mechanism. It projects input features into a new space. Then the attention weights are decided by inner products between features in this new space. We normalize the attention matrix so that the diagonal elements of $\bU$ are always one. 
     
In each layer $\ell$, we get an attention matrix $\bU_\ell$ as above. Then we use $\bA_{\ell}^{att} = \bA \odot \bU_\ell$ as the new 
adjacency matrix used in layer $\ell$. The actual computation is 
\begin{align}
\bH^{\ell} &= \sigma\left(\bA_{\ell}^{att} \bH^{\ell - 1} \bW^{\ell} \right), ~~~ \ell = 1, \ldots, L \label{eq:kcnatt}
\end{align}

We call this new model the KCN-att. When the matrix $\bW_{att}$ has small weights, then $\bU$ approaches a matrix with all entries 
being one. In this case, the KCN-att becomes similar to the KCN.  When the matrix $\bW_{att}$ has large weights, then $\bU$ tends to approach the identity matrix, and then the KCN-att tends to reduce neighbors' influence.

\subsection{KCN based GraphSAGE}

We also use GraphSAGE \citep{hamilton2017inductive} as the predictive model of the KCN given that GraphSAGE performs well on several node classification tasks. GraphSAGE cannot use a weighted graph, so we treat the graph over the neighborhood of $i$ as a complete graph. Let $\bH^{\ell-1} = \{\bh^{\ell-1}_k: k \in \beta_i\}$ be the input to the GraphSAGE layer $\ell$, then the layer computes its output $\bH^{\ell}$ as follows.   
\begin{align}
\bg^{\ell}_j &= \mathrm{AGG}\left(\{\bh^{\ell - 1}_k,  k \in \beta_i, k \neq j \}\right),   ~~~~\forall j \in \beta_i \\
\bh^{\ell}_j &= \sigma\left(\bW^{\ell}_1 \bh_j^{\ell - 1}  +  \bW^{\ell}_2 \bg_j^{\ell} \right),   ~~~~ \forall j \in \beta_i\\
\bH^{\ell} &= \left\{ \bh^{\ell}_j / \|\bh^{\ell}_j\|_2:   \forall j \in \beta_i \right\}
\end{align}
The function $\mathrm{AGG}(\cdot)$ aggregates a list of vectors into one. We use the max-pooling aggregator, one the three aggregators proposed in the original work \citep{hamilton2017inductive}. 
\begin{align*}
\mathrm{AGG}(\bH^{\ell - 1}_{\beta_i \backslash j}) = \max({\sigma(\bW_{pool} \bh^{\ell-1}_k + \bb), k \in \beta_i, k\neq j})
\end{align*}
Here $\max$ takes the element-wise max values over a list of vectors. We refer to this model as the KCN-sage.


\section{Analysis}

\subsection{Computation Complexity}

The time complexity of KCN and the two variants includes nearest-neighbor search and network training. In order to find the $K$ nearest neighbors we utilize a KD tree, which takes $O(N\log(N))$ time to build. Here we treat the dimensionality of spatial coordinates as a constant because it usually a small number (2 or 3). Querying a single data point in the tree takes time $O(K\log(N))$, and searching neighbors for all data points takes a total of $O(NK\log(N))$ time. 

When we train the model on a single instance, the computation of the adjacency matrix takes time $O(K^2)$. The computation within each layer takes time $O(K^2 d_{\mathrm{max}})$, with $d_{\mathrm{max}}$ being the largest dimensionality of hidden layers. The forward computation and backpropagation for one instance takes time  $O(K^2 L d_{\mathrm{max}})$, and one training epoch takes time $O(N K^2 L d_{\mathrm{max}})$. 

\subsection{Relation to Kriging}

KCN is a flexible model and approximately includes local kriging as a special case. This fact is shown by the following theorem.   

\noindent\textbf{Theorem 1:} Assume the variogram of a kriging model satisfies $2\gamma(0) > 0$ \footnote{The value $2\gamma(0)$ is called the nugget of the variogram, which is usually greater than zero.}. Also assume $\tilde{\bX} = [\bx_*, \bX_{\alpha_*}^\top]^\top$ has full column rank. Then there exists a set of special parameters and activations with which a KCN makes the same prediction as the kriging prediction, i.e. $\hat{y}_*^{KCN} = \hat{y}_*^{kriging}$. 

\noindent\textbf{Proof sketch:}
Let $\tilde{\bGamma}$ be the covaraince matrix corresponding to the new data point and training data point. 
\begin{align} 
\tilde{\bGamma} =
\left[\begin{array}{cc}
0 & \bgamma^\top \\ 
\bgamma & \bGamma
\end{array} \right].
\label{eq:Gamma-tilde}
\end{align}
Here $\bgamma$ and $\bGamma$ are semivariograms defined in the same way as kriging. 
To approximate kriging, we set the KCN to have one convolutional layer 
and a dense layer. We set 
\begin{align}
\bar{\bA} = \tilde{\bGamma}^{-1} + \tilde{\bGamma}^{-1} \tilde{\bX} ( \tilde{\bX}^{\top } \tilde{\bGamma}^{-1}\tilde{\bX})^{-1} \tilde{\bX}^{\top} \tilde{\bGamma}^{-1} 
\label{eq:A-tilde}
\end{align}
as the ``normalized adjacency matrix'' and directly use it to multiply the hidden input. We consider a 1-layer GCN with a special activation function $\sigma_{div}(\cdot)$. The first row of the GCN output is $\be^\top \bH^{L} = \sigma_{div}\left(\be \bar{\bA} \bH^0 \bW^1\right)$. In the Appendix we show $\be\bar{\bA}  = \left[z^{-1}, - z^{-1} \blambda^\top \right]$, then  
\begin{align*}
\be\bar{\bA} \bH^0 = \left[- z^{-1} \blambda^\top \by_{\alpha_*}, ~z^{-1}, ~z^{-1} (\bx_* - \blambda^{\top} \bX_{\alpha_*}) \right].
\end{align*}
Here $z$ is a scalar, and $\blambda$ is the kriging coefficient defined in \eqref{eq:lambda}. Let $\bW^1$ be the matrix taking the first two elements of the vector, then $\be\bar{\bA} \bH^0 \bW^1 = [-z^{-1} \blambda^\top \by_{\alpha_*}, z^{-1}]$. Denote it as $\bu$.  
Define the activation function to be $\sigma_{div}(\bu) = [- u_1 / u_2, 0]$, set $\bw_{den} = [1, 0]^\top$, and set the activation of the fully connected layer to be identity, then the KCN predicts $\hat{y}_*^{KCN} = \blambda^\top \by_{\alpha_*}$, which is exactly the same as  kriging prediction $\hat{y}_*^{kriging}$. 

In the real implementation, we use normal activation functions such as ReLU. The combination of the first two rows of $\bW^1$, the GCN activation, and the dense layer can be viewed as two-layer feedforward network applied to $\bu$. If the two-layer neural network can emulate the function $- u_1 / u_2$, then a normal setting of the network can also approximate kriging well. \hfill $\square$

This theorem and its proof have strong implications for our model development. First, if the KCN uses $\bar{\bA}$ defined above as the normalized Laplacian, then the KCN has a straightforward way to discover kriging solutions. Since $\bar{\bA}$ has a small size,  $(K+1) \times (K+1)$, the computation of $\bar{\bA}$ is affordable. Second, the matrix  $\bar{\bA}$ indicates that we should introduce the feature matrix into the computation of the ``normalized Laplacian''. Otherwise, the KCN may need complicated computations to recover kriging results. This is one main motivation behind our usage of graph attention in KCN-att.


\section{Experiment}

We evaluate our methods on three tasks: bird count modeling, restaurant rating regression, and precipitation regression. We use Kriging, Random Forest, Graph Convolution Network, and Graph SAmple and aggreGatE as baselines.

\subsection{Experiment setup}

\begin{table*}[t]
    \centering
   \begin{tabular}{ |c||cccc|ccc| } 
   \hline
   methods & Kriging & RF & GCN & GraphSAGE & KCN & KCN-att & KCN-sage  \\ 
   \hline
   MSE   & 1.56 $\pm$ .85 & 0.68 $\pm$ .03 & 0.70 $\pm$ .02 & 0.53 $\pm$ .01 & 0.50 $\pm$ .01  & 0.49 $\pm$ .01 & \textbf{0.44 $\pm$ .01} \\
   \hline
   NLL  &  n.a. & n.a. & 1.82 $\pm$ .00 & 1.73 $\pm$ .00 & 1.60 $\pm$ .00  & 1.58 $\pm$ .00 & \textbf{1.51 $\pm$ .00} \\
   \hline
   \end{tabular}
    \caption{Experiment results on the bird count dataset. Performances are measured by the mean squared error and the negative log likelihood of preditions. Smaller values are better.}
    \label{tab:birdperf}
\end{table*}

\noindent \textbf{Kriging}: we use the implementation of Kriging within Automap \citep{automap}. Automap essentially automates the process of Kriging, by automatically fitting variograms, and testing several different models. In all of our experiments Automap tests spherical, exponential, gaussian, matern, and stein variograms and picks the best one based on the smallest residual sum of squares. Since all of the datasets have a large number of data points, we use local kriging and only consider the closest 100 points.

\noindent \textbf{Random Forest}: \citet{hengl2018random} use Random Forest to make predictions for spatial data. For each data point, the algorithm calculates the distances between that point and all training points. These distances are then used as the feature vector of that data point. This algorithm does not scale to very large datasets, so we downsample the training set to a size of 1000. We use the implementation of Random Forest \citep{ranger}, and method of tuning \citep{tuneRanger} used by the authors of \citet{hengl2018random}. The implementation tunes four hyperparameters of Random Forest: the number of trees to use, the number of variables to consider at a node split, the minimal node size, and the sample fraction when training each tree.

\noindent \textbf{GCN}: we modify Kipf's implementation of \citep{kipf2016} for regression problems. Before we run the GCN on spatial data, we first build a undirected graph over data points: we connect two data points if one is among the other's $K$ nearest neighbors. We only consider a GCN with two hidden layers. We tune the hyper-parameters of the GCN in the same way as we tune the KCN and the KCN-att below. 

\noindent \textbf{GraphSAGE}: we implement GraphSAGE with the \textit{Spektral} graph deep learning library. For each experiment, we build an undirected graph in the same way as the GCN. Then we train a two hidden layer GraphSAGE, with hyperparameters are tuned as below.

\noindent \textbf{KCN  \& KCN-att \& KCN-sage}: the three models use two hidden layers respectively. We tune the following hyperparameters: hidden sizes $\in ((20, 10), (10, 5), (5, 3))$, dropout rate $\in (0, 0.25, 0.5)$, and kernel length $\in (1, .5, .1, .05)$. Note that GraphSAGE and KCN-sage do not consider weighted adjacency matrix, so there is no need to tune kernel length for them. We also employed early stopping to decide the number epochs. 

\subsection{Bird count modeling}
\begin{figure}[t]
\centering
    \begin{tabular}{cc}
    {\raisebox{0.15 in}{\includegraphics[width=.35\linewidth]{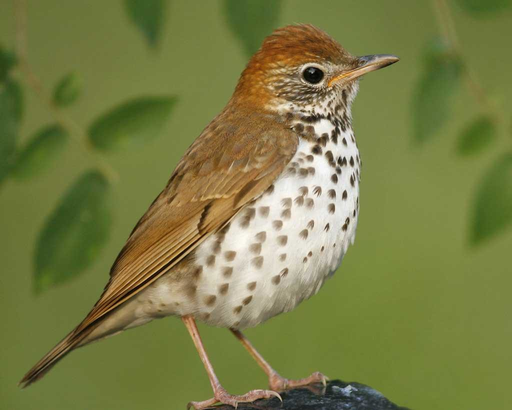}}}
    \hfill&  \hfill
    {\includegraphics[width=0.4\linewidth]{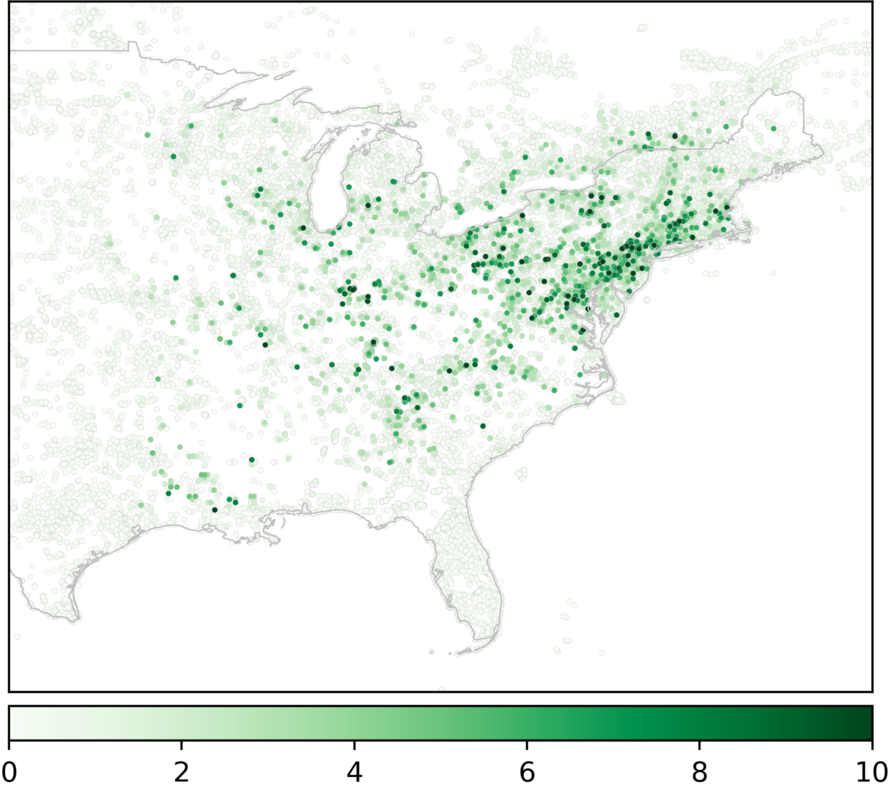}}
    \\
    (a) & (b)
    \end{tabular}
\caption{Wood thrush (a) and observed counts over eastern US, June 2014 (b). }
\label{fig:birds_dist_and_picture}
\end{figure}
One application of our KCN models is modeling bird count data from the eBird project \citep{fink2010}, which contains over one billion of records of bird observation events. Modeling bird data from the eBird project provides an opportunity to deepen our understanding of birds as part of the ecosystem. In this experiment, we model the distribution of \textit{wood thrush} in June, which is of great interests to onithologists \citep{johnston2019best}. Figure \ref{fig:birds_dist_and_picture} shows a picture of a wood thrush and the distribution of observed counts over the eastern US.  

We restrict our data to a subset of records of wood thrush in June 2014. Each record has a GPS location, a count of wood thrushes observed, and a list of features such as observation time, count type (stationary count, traveling count, etc.), effort hours, and effort area. After removing 583 records with uncertain counts or counts over 10, we get 107,246 records to form our dataset. Bird counts in this dataset are highly sparse: only 11,468 records (fraction of 0.11) have positive counts. We split the dataset into a training set and a test set by 1:1.

\begin{figure}[t]
    \centering
    \includegraphics[width=0.35\textwidth]{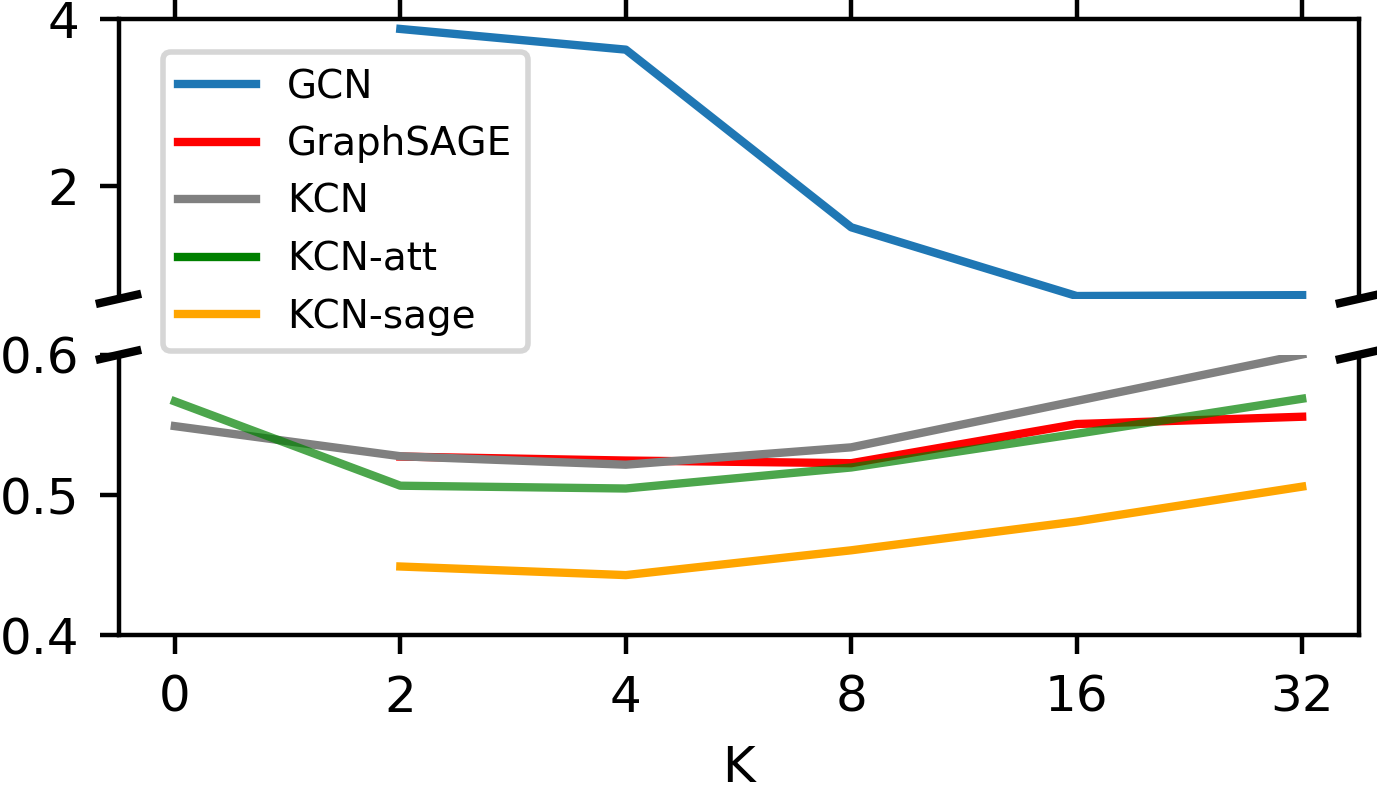}
    \caption{Mean Squared Error of GCN, GraphSAGE, KCN, KCN-att, and KCN-sage using different numbers of neighbors.}
    \label{fig:num_nb}
\end{figure}

When we test our models and baselines, we consider two evaluation metrics. The first one is mean squared error (MSE), so we have a fair comparison with Kriging, the minimization objective of which is the mean squared error. The second one is negative log-likelihood. We use a zero-inflated Poisson distribution \citep{lambert92} as the predictive distribution for each count. The model needs to output a logit $u$ for the Bernoulli probability and the mean $\lambda$ of the Poisson component. The probability of a count $y$ given $u$ and $\lambda$ is
\begin{align}
p(y) = \left\{
\begin{array}{ll}
(1 - \mathrm{expit}(u)) + p_{poisson}(y=0) & \mbox{ if } y=0, \\
p_{poisson}(y=0) \mbox{ if } & y > 0. 
\end{array}\right.
\end{align}

\begin{table*}[t!]
    \begin{center}
            \begin{tabular}{  |c||cccc|ccc|  } 
                \hline
                Method & Kriging & Random Forest & GCN & GraphSAGE & KCN & KCN-att & KCN-sage\\ 
                \hline
                MSE & 1.49 $\pm$ .008 & 1.04 $\pm$  .005 & 1.37 $\pm$ .006 & \textbf{0.969 $\pm$ .004} & 0.990 $\pm$ .004 & 0.977 $\pm$ .004 & \textbf{0.959 $\pm$ .004} \\ 
                \hline
            \end{tabular} 
        \caption{The results on the dataset of restaurant ratings. Performances are measured by MSE. Smaller values are better. }
        \label{tab:YelpResults}
    \end{center}
\end{table*}%

\begin{table*}[t!]
    \begin{center}
        \begin{tabular}{ |c||cccc|ccc| } 
            \hline
            Method & Kriging & Random Forest & GCN & GraphSAGE & KCN & KCN-att & KCN-sage \\ 
            \hline
            MSE & .155 $\pm$ .013 & .046 $\pm$ .003 & .640 $\pm$ .023 & .056 $\pm$ .003 &  \textbf{.029 $\pm$ .002} &  \textbf{.029 $\pm$ .002} & \textbf{.030 $\pm$ .002} \\ 
            \hline
        \end{tabular} %
    \caption{Experiment results on the precipitation dataset. Performances are measured by MSE. Smaller values are better.}
    \label{tab:PrecipResults}
    \end{center}
\end{table*}

Table \ref{tab:birdperf} shows the performance of KCN, KCN-att, KCN-sage, and baseline models. From this table, we can see that the three KCN models significantly outperform baseline methods. We also observed that GraphSAGE based methods have superior performance than the GCN based methods, we speculate it is because of the concatenation operation (plays a role similar to a skip link) used in the GraphSAGE. Given that bird counts are highly non-Gaussian, we don't expect kriging to perform very well. Random Forest gets much better performance than kriging, but it overly smooths the training data given the small number of training points it can use. The KCN and the KCN-att achieve similar performances.

We also study the performances of the GCN, GraphSAGE, KCN, KCN-att, and KCN-sage when different numbers of neighbors are used to form the graph.
Figure \ref{fig:num_nb} shows performance values of the five models using different numbers. The GCN perform poorly when the number of neighbors is small in the construction of the graph. In this case, a test point might only connect to another test data point, then the message propagation between two test points is not helpful. GraphSAGE is robust to the number of neighbors. In the KCN models, a data point has $K$ training points as direct neighbors, so KCN models can make better use of the training data in this sense. When a KCN models uses zero neighbors, it is equivalent to a fully connected neural network, and its performance deteriorates significantly. It indicates that spatial correlation exists in the data. KCN, KCN-att, and KCN-sage only need a small number of neighbors to perform well. We speculate that a bird or its nest can be observed only in a small spatial range, so the correlation between near sites are strong but diminishes quickly as the distance increases. The KCN-att performs slightly better than the KCN because the KCN-att is able to use observatory features to decide whether a neighboring count is from the same situation or not.

\subsection{Restaurant rating regression}

Yelp is a popular rating website, which allows users to rate and provide information about businesses. They have hosted a large collection of these business ratings and attributes for download. In this experiment, we only consider the restaurants within that dataset. Each restaurant has a GPS location and an average rating rounded to the nearest .5, from 0 to 5. Additionally, we choose 13 related attributes from the dataset, all but one of which is categorical. We turn these categorical covariates into 30 indicator variables. These indicators give information about restaurant attributes such as whether it serves alcohol, and whether or not it takes credit card. After we drop any rows where the ratings, coordinates, or number of reviews is NA, we obtain 188,586 restaurants. We then split the data 1:1 to form a training and test set.

Table \ref{tab:YelpResults} shows the experiment results on this dataset. The KCN, the KCN-att, and the KCN-sage improve the performance of their corresponding baseline models. The KCN, the KCN-att, the KCN-sage, and the GraphSAGE outperform baseline models by a small margin. The regression task on this dataset is a hard one. The features seem to not be very useful. It is actually hard to overfit the labels with a normal feedforward neural network. However, there are some weak spatial effect. The average rating over the entire dataset achieves a mean squared error of $1.01$ while the average of the nearest $35$ neighbors results in a mean squared error of $.984$. This is understandable, since it is normal that good restaurants and bad restaurants mix in the area. In this experiment, we find that Kriging is very stable when features are discrete and sparse. We add a small amount of noise to the feature matrix to avoid numerical issues.

\subsection{Precipitation regression}

\begin{figure}[t]
\centering
    \includegraphics[width=.36\textwidth]{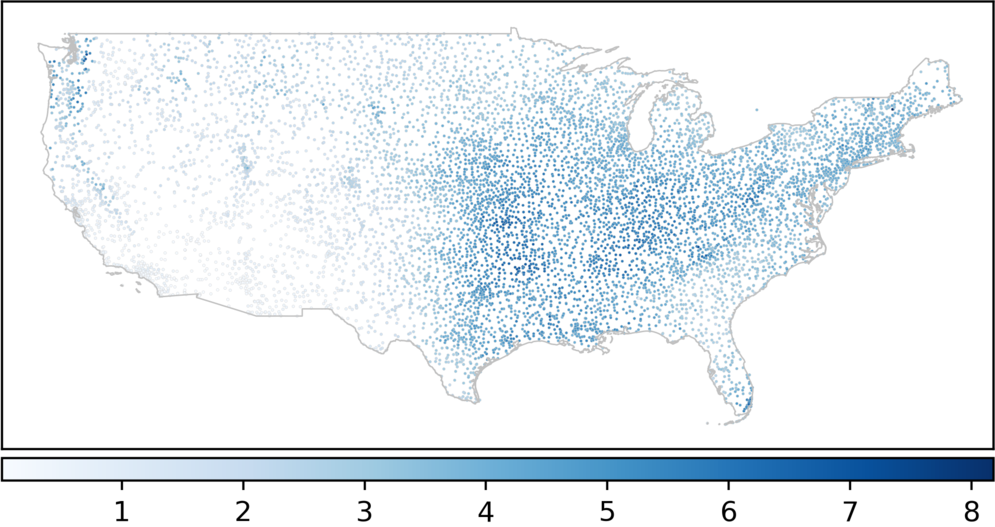}
    \caption{Distribution of Precipitation values.}
    \label{fig:precip_dist}
\end{figure}

The National Oceanic and Atmospheric Administration keeps detailed records of precipitation levels across the United States. One such dataset provides monthly average precipitation in inches from 1981 to 2010 across the US. We average the precipitation level in May for 8,832 stations. We then take the log of these average precipitation values as target values for the regression task. Essentially, we assumes a log-normal distribution of precipitation levels. Finally we have a target value, coordinates of each station, and one feature (the elevation) of each station. Data are split with portion 1:1 as a training and testing. Figure \ref{fig:precip_dist} shows the data distribution over the US. 

Table \ref{tab:PrecipResults} summarizes the experimental results using the mean squared error. The target values in the log-scale are more likely to be from a Gaussian distribution than the previous two datasets, so Kriging performs relatively well compared to other methods. The Random Forest method only uses 1000 data points as the training data, so it omits a lot of detailed variations. The GCN models perform poorly on this dataset. One reason is that there are not many features for the GCN to learn. The GCN model becomes more like a ``generative'' model that generates observations from hidden values. Compared to the GCN, the KCN models particularly benefit from this dataset because the KCN models work more like a discriminative model. Note that discriminative models often outperform generative models in supervised learning tasks.  


\section{Conclusion}

In this work, we introduced the Kriging Convolutional Network, a novel approach to modeling spatial data. Like kriging, the KCN model directly use training labels in the prediction. However, it enjoys the flexibility of neural networks by using GNNs as the backbone. 
We further introduce the attention mechanism to the model to create the KCN-att model.  The KCN-att model has better control over which neighbors to use. Our analysis also reveals that KCN has a straightforward method to approximate kriging models. In the empirical study, we have compared KCN and KCN-att with three baselines on three applications. The experiment results shows 
the superiority of two KCN models over baselines. They indicates that feeding in observed labels to the model is a powerful way to improve the performance. 


\section{Appendix: Detailed Proof of Theorem 1}

We need to derive $\bu = \be \bar{\bA} \tilde{\bX}$, where $\be = [1, \bzero^\top]^\top$, $\bar{\bA}$ is ``normalized adjacency matrix'' defined in \eqref{eq:A-tilde}, and $\tilde{\bX} = [\bx_*, \bX^\top]^\top$ are feature vectors.
We create the following shorthand notations to facilitate our derivation. 
\begin{align*}
& t = - \bgamma^\top \bGamma^{-1} \bgamma, ~~ 
\ba = -\bGamma^{-1} \bgamma, ~~ 
\bc = \bx_* + \bX^{\top}\ba, \\
& \bT = \bX^{\top} \bGamma^{-1} \bX, ~~  
\bB = \bX \bT^{-1}, ~~  
r = \bc^\top \bT^{-1} \bc 
\end{align*}

We will show that  the first row of $\bar{\bA}$ is 
\begin{align*}
&\be^\top \bar{\bA} = \left[z^{-1}, -z^{-1} \blambda^{\top} \right], \\
&z = - \bgamma^\top \bGamma^{-1} \bgamma + \bgamma^\top \bGamma^{-1} \bX \bT^{-1} \bX^\top \bGamma^{-1} \bgamma.
\end{align*}

By checking \eqref{eq:A-tilde}, we first compute the inverse $\tilde{\bGamma}^{-1}$ is
\begin{align*}
\tilde{\bGamma}^{-1} &= 
\left[\begin{array}{cc}
t^{-1} & t^{-1}\ba^{\top} \\
t^{-1}\ba & \bGamma^{-1} + t^{-1} \ba \ba^{\top}
\end{array}\right] \\
&= 
 t^{-1}  
\left[\begin{array}{c}
1 \\ 
\ba
\end{array} \right]
[1, \ba^{\top}] 
+
 \left[\begin{array}{cc}
0 & \bzero^\top \\
\bzero & \bGamma^{-1} 
\end{array}\right].
\end{align*}
Denote $\bv_1 = \be \tilde{\bGamma}^{-1} = t^{-1}[1, \ba^{\top}]$.

We then consider the second term in \eqref{eq:A-tilde}. We have 
\begin{align*}
\tilde{\bX}^{\top} \tilde{\bGamma}^{-1} &= t^{-1} \bc [1, \ba^{\top}] + [\bzero, \bX^{\top}\bGamma^{-1}] \\
& = [ t^{-1} \bc,  t^{-1} \bc \ba^{\top} +  \bX^{\top}\bGamma^{-1}]. 
\end{align*}
Denote $\bS = (\tilde{\bX}^\top \tilde{\bGamma}^{-1} \tilde{\bX})^{-1}$,   
\begin{align*}
\bS &= (t^{-1} \bc \bc^{\top} + \bT)^{-1}  \\
&= \bT^{-1} - \bT^{-1} \bc (t + \bc^\top \bT^{-1} \bc)^{-1} \bc^{\top} \bT^{-1}.
\end{align*}
The first line is from the equation $[1, \ba^\top] \tilde{\bX} = \bc^{\top}$.

Denote $\bv_2 = \be^\top \tilde{\bGamma}^{-1} \tilde{\bX} \bS \tilde{\bX}^{\top} \tilde{\bGamma}^{-1}$. Insert 
the expansion of $\tilde{\bX}^{\top} \tilde{\bGamma}^{-1}$, we have
\begin{align*}
\bv_2 &= t^{-1} \bc^{\top} \bS [t^{-1} \bc, ~~ t^{-1} \bc \ba^{\top} + \bX^{\top}\bGamma^{-1}].
\end{align*}

By $\bc^\top \bS = t(t+r)^{-1} \bc \bT^{-1}$ and $w = \bc^\top \bS \bc = rt(t+r)^{-1}$, we have 
\begin{align*}
\bv_2 &= (t+r)^{-1}[r t^{-1}, ~ r t^{-1} \ba^{\top} + \bc^{\top} \bT^{-1} \bX^{\top}\bGamma^{-1}]. 
\end{align*}

Since $t^{-1} - (t + r)^{-1}rt^{-1} = (t + r)^{-1}$, we have
\begin{align*}
\be^\top \bar{\bA} &= \bv_1 - \bv_2\\
&= (t + r)^{-1} [1, ~~ (\ba^{\top} - \bc^\top \bT^{-1} \bX^{\top}\bGamma^{-1})]. 
\end{align*}
Then we expand $\ba$ and $\bc$ to get 
\begin{align*}
\be^\top \bar{\bA} &= (t + r)^{-1} [1, ~~ -( \bgamma + \bB \bx_* - \bB\bX\bGamma^{-1}\bgamma)^\top\bGamma^{-1})] \\
 &= z^{-1} [1, ~~ - \blambda^\top] 
\end{align*}
Here $z = t + r$. 

Since $\bar{\bA}$ are computed from normal matrix operations, its entries are bounded. Therefore, $z \neq 0$.  

\section{ Acknowledgments}
This work has been supported in part by Gordon \& Betty Moore Foundation, NSF CISE-1908617, and and NSF CRII-1850358. We thank all reviewers of this work for their insightful comments.

\bibliography{ref-liping}
\bibliographystyle{aaai}

\end{document}